# A 8 bits Pipeline Analog to Digital Converter Design for High Speed Camera Application

**Eri Prasetyo, Hamzah Afandi, Nurul Huda )***
**Dominique Ginhac , Michel Paindavoine )****

*Center for Microelectronics & Images Processing*
*Gundarama university , Indonesia*

**Laboratoire LE2I - UMR CNRS 5158*
*Université de Bourgogne*
*21078 Dijon Cedex – FRANCE*

E-mail: (dginhac,paindav)@u-bourgogne.fr
(eri, huda,hamzah)@staff.gunadarma.ac.id

*Abstract -* **This paper describes a pipeline analog-to-digital converter is implemented for high speed camera. In the pipeline ADC design, prime factor is designing operational amplifier with high gain so ADC have been high speed. The other advantage of pipeline is simple on concept, easy to implement in layout and have flexibility to increase speed. We made design and simulation using Mentor Graphics Software with 0.6 µm CMOS technology with a total power dissipation of 75.47 mW. Circuit techniques used include a precise comparator, operational amplifier and clock management. A switched capacitor is used to sample and multiplying at each stage. Simulation a worst case DNL and INL of 0.75 LSB. The design operates at 5 V dc.
The ADC achieves a SNDR of 44.86 dB.**
*keywords: pipeline, switched capacitor, clock management*

# A 8 bits Pipeline Analog to Digital Converter Design for High Speed Camera Application

*Abstarct* - **This paper describes a pipeline analog-to-digital converter is implemented for high speed camera. In the pipeline ADC design, prime factor is designing operational amplifier with high gain so ADC have been high speed. The other advantage of pipeline is simple on concept, easy to implement in layout and have flexibility to increase speed. We made design and simulation using Mentor Graphics Software with 0.6 µm CMOS technology with a total power dissipation of 75.47 mW. Circuit techniques used include a precise comparator, operational amplifier and clock management. A switched capacitor is used to sample and multiplying at each stage. Simulation a worst case DNL and INL of 0.75 LSB. The design operates at 5 V dc.**
**The ADC achieves a SNDR of 44.86 dB.**

## 1. Introduction

CMOS image sensors have evolved in the past years as a promising alternative to the conventional Charge Coupled device (CCD) technology. CMOS offer lower power consumption, more functionality and the possibility to integrate a complete camera system on one CHIP.

The high speed camera used matrices photodiode to capture objects and each photodiode send an analog pixel to matrices column. Output analog pixel is converted to digital pixel by ADC then output from ADC is processed by digital processor element. ADC is used to converter is pipeline. Diagram block high speed camera is shown in figure 1.

In the real time images processing, sensors function is important because it has function as transducer, so images can be processed to application for examples, face tracking and face recognition, medical imaging, industrial, sports and so on[4,5].

Figure 1. Describes 64x64 active pixel sensors (APS) is used capture object. We used the row decoder is charged to send to each line of pixels the control signals. The automatic scan of the whole array of pixels or sub windows of pixels is implemented by a sequential control unit which generates the internal signals to row and column decoders.

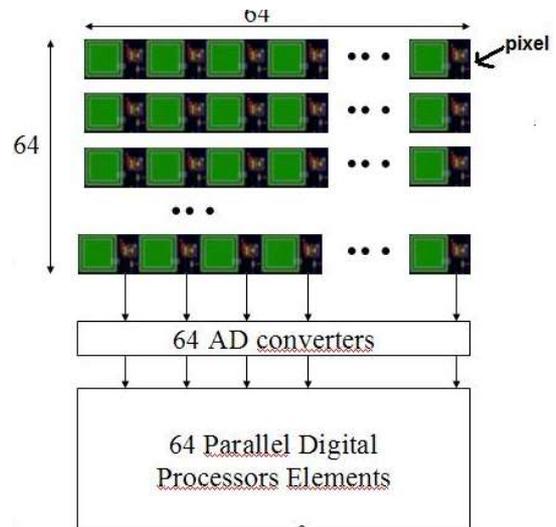

Figure 1. Diagram Block high speed camera

Number of ADC is used in the system are 64 on parallel condition. Function of ADC in the process is important to convert from analog pixels to digital pixels where output from APS is nearly 4K pixel with each pixel < 100 ns or same as 400 µS per images or 2500 images/s. so wherever we must design pipeline ADC which have transfer rate 80 Msamples/s.

## 2. One-Bit Per Stage Pipeline Architecture

Figure 2. One bit/ stage architecture

Figure 2 shows block diagram of an ideal N-stage, 1-bit per stage pipelined A/D converter. Each stage contributes a single bit to digital output. The most significant bits are resolved by first stage in the pipeline. The result of stage is passed on the next stage where the cycle is repeated. A pipeline stage is implemented by the conventional switched capacitor (SC), it is shown at figure 3[4].

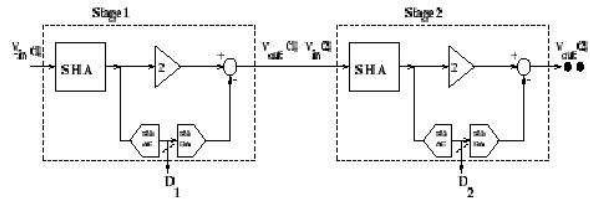

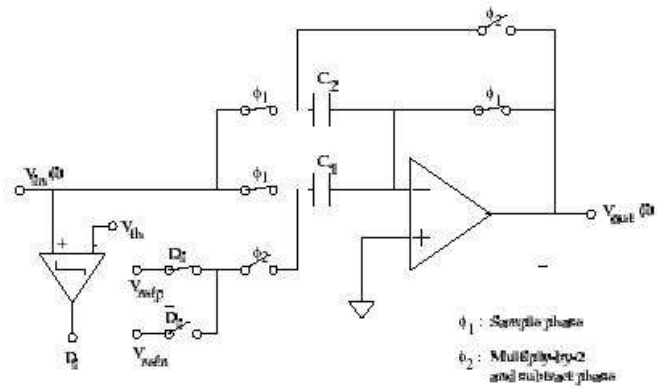

Fiure. 3. Scheme of switched capacitor pipelined A/D converter

V*refp* is the positive reference voltage and V*refn* is a negative reference voltage. Each stage consists of capacitor C1, C2, an operational amplifier and a comparator. Value of *C1* and *C2* are equal in my design. Each stage operates in two phases, a sampling phase and a multiplying phase.

During the sampling phase $\phi 1$, the comparator produces a digital output *Di*. *Di* is 1 if *Vin* > *Vth* and *Di* is 0 if *Vin* < *Vth*, where *Vth* is the threshold voltage defined midway between *Vrefp* and *Vrefn*. During multiplying phase, C2 is connected to the output of the operational amplifier and C1 is connected to either the reference voltage *Vrefp* or *Vrefn*, depending on the bit value *Di*. If *Di* = 1, C1 is connected to *Vrefp*, resulting in the resedu ( Vout ) is :

$$Vout(i) = 2 \ x \ Vin(i) – Di.Vrefp \tag{1}$$

Otherwise, C1 is connected to *Vrefn*, giving an output voltage :

$$Vout(i) = 2 \ x \ Vin(i) - \check{D}.Vrefn \tag{2}$$

## 3. Comparator

Precision comparator is implemented to each stage of the ADC. We prefer to use precision comparator then digital correction to minimize offset error of comparator and better output of ADC.

This comparator consists of three blocks: preamplifier, decision circuit and output buffer. First block is the input preamplifier which the circuit is a differential amplifier with active loads. The size of transistors m_2 and m_3 are set by considering the diff-amp transconductance and the input capacitance. Second block is a positive feedback or decision circuit, it is the heart of the comparator. The circuit uses positive feedback from the cross gate connection of m_11 and m_12 to increase the gain of the decision element. Third stage is output buffer; it is to convert the output of the decision circuit into a logic signal. The inverter (m_20 and m_21) is added to isolate any load capacitance from the self biasing differential amplifier. The complete circuit of comparator is shown in figure 4.

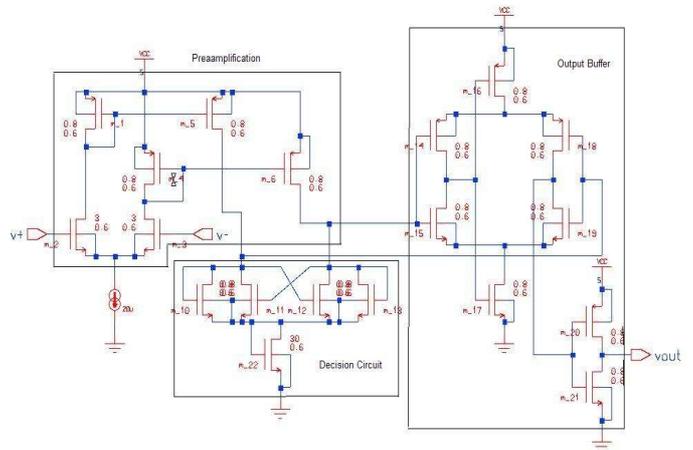

Figure 4. The comparator circuit

## 4. Operational Amplifier

In this pipeline ADCs, operational amplifier is very important to get accurately result. We used an operational transconductance amplifier which has a gain of approximately 55 dB for a bias current of 2.5 μA with Vdd = 5 V and Vss = -5 V. A value of loading capacitor is 0.1 Pf. The complete circuit is shown in figure 5. Transistors m_1_1_1 and m_1_1 functions as a constant current source, and transistors m_1, m_2 and m_3 functions as two current mirror 'pairs'. The transistors m_4, m_5, m_6 and m_7 are the differential amplifier.

Transistor m_9 is an output amplifier stage. In the simulation, we got the resultat for phase margin (PM) was -145 degre, A gain was 55 dB and Gain bandwidth product was 800 MHz. A power dissipation mesured of 10.825 mW.

## 5. Clock Management

In the design pipeline A/D converter use latch technique is used to hold active condition at multiplying ø2 (phi2) and non active condition at sampling ø1 (phi1) until next stage begin to execute sampling phase. This purpose to keep the output voltage of residu from before stage conformity at input next stage.

The clock management system use counter to count some clock to active the address decoder from each stage.

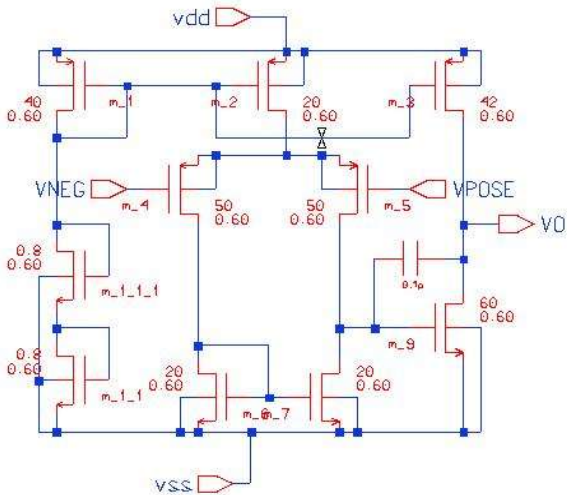

Figure 5. Transconductance OP-AMP

Signal output decoder active reset signal so the clock management begin working. This work is begun from early address to last address. Ending of address decoder, a stop decoder give reset signal
Stopping activity pipeline ADC's. The complete circuit is shown at figure 6.

Figure 6. The circuit of clock management

## 6. Result

One stage A/D converter layout was estimated to occupy about 174 μm x 89 μm, it is seen at figure 7.

Figure 8 shows the dc linearity of the ADC at conversion rate of 20 Msamples/s. In the figure 8(a), the CODE is plotted versus integral nonlinearity (INL) value and figure 8(b), the CODE is plotted versus differential nonlinearity (DNL). Note that since each simulation lasted 20 minutes, only 25 codes were tested. As shown, the worst INL is less than 0.8 LSB; the DNL is less then 0.8 LSB.

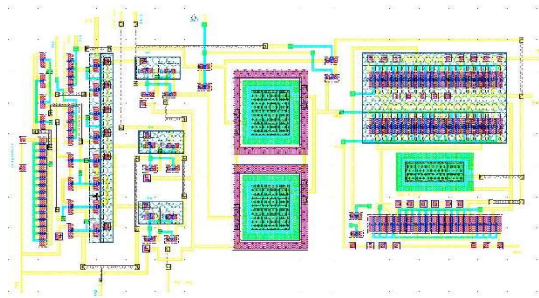

Figure 7. One stage A/D converter layout

Figure 9 shows the output of Fast Fourier transform (FFT) on a block of 1024 consecutive codes. The conversion rate is 20 Msamples/s, and the input is full scale sines wive at 10 MHz. From curve FFT, The signal-to-noise plus distortions ratio (SNDR) is obtained about 44.86 dB. The effective number of bits (ENOB) is calculated environ 7.2 bits.

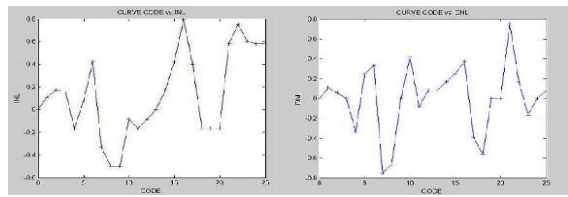

Figure 8. (a) Curve of Code vs INL and (b) Curve of Code vs DNL

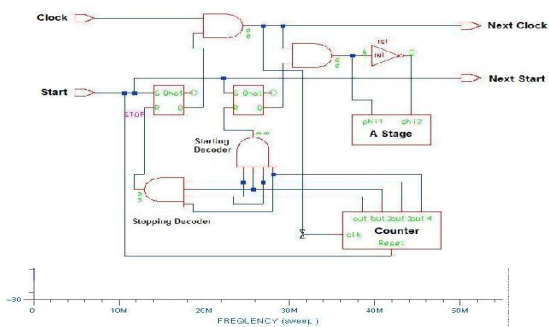
Figure 9. Curve of FFT

## 7. Conclusion

The pipeline ADC 8 bits, 80 Msamples/s were implemented in 0.6 μm technology with total power dissipation 75.47 mW. Refer to result of experiment; the ADC can be implemented for high speed camera.

The system use clock management to manage data conversion so that the system is Simple and have good precision.